\documentclass[cameraready]{Interspeech}
\usepackage{subcaption}
\usepackage{color}
\usepackage{tabularray}
\definecolor{Silver}{rgb}{0.8,0.8,0.8}

\title{Domain-Aware Mispronunciation Detection and Diagnosis Using Language-Specific Statistical Graphs}

\author[affiliation={1,2}, equalcontribution]{Huu Tuong}{Tu}

\author[affiliation={3}, equalcontribution]{Hanh}{Nguyen}

\author[affiliation={3}]{Thien Van}{Luong}

\author[affiliation={2}]{Nguyen Tien}{Cuong}

\author[affiliation={3}, correspondingauthor]{Vu}{Huan}

\author[affiliation={1}, correspondingauthor]{Nguyen Thi Thu}{Trang}

\address{
    $^1$Hanoi University of Science and Technology \quad
    $^2$VNPT AI, VNPT Group \\
    $^3$National Economics University
}


\usepackage{comment}

\begin{document}

\maketitle

\begin{abstract}
Mispronunciation Detection and Diagnosis (MDD) has gained increasing importance in computer-assisted language learning and speech technology in recent years. In this paper, we propose a method for constructing statistical graphs that enable models to learn phoneme confusion patterns represented as directed graphs. Furthermore, we introduce a language-specific strategy to capture systematic pronunciation differences across various native language (L1) backgrounds. The effectiveness of our approach is demonstrated through extensive experiments on the L2-ARCTIC benchmark, where it achieves an F1-score of 59.52\%, outperforming several competitive baselines.
\end{abstract}

\section{Introduction}

Mispronunciation Detection and Diagnosis is a major component of Computer-Assisted Pronunciation Training (CAPT) systems, detecting pronunciation errors and providing diagnosis or personalized feedback to language learners. With the help of MDD, learners can improve their foreign language acquisition and overall proficiency. 

Early mispronunciation detection research was primarily grounded in pronunciation scoring approaches, which use the Goodness of Pronunciation (GOP) \cite{WITT200095, gop1, gop2}, applied to the acoustic model, to estimate phone-level confidence scores via forced alignment, forming a practical basis for error detection with thresholding. Beyond that, rule-driven methods were introduced, most notably via Extended Recognition Networks (ERN) \cite{harrison09_slate, 7415299, ern1}, where phonological rules are compiled into recognition networks to represent common L2 error patterns. This helps not only to assess learners’ pronunciation but also provide detailed feedback.

In recent years, MDD has shifted towards neural and end-to-end (E2E) paradigms. Data-driven models began to learn phonological relationships between acoustics and canonical pronunciations \cite{7752846}, and later E2E architectures such as CNN-RNN-CTC \cite{8682654} constructed sequences of phonemes based on the acoustic data in the CTC \cite{ctc} model and performed MDD by aligning the sequence with canonical phonemes. Furthermore, text-dependent MDD models were introduced to improve MDD performance by leveraging prior text information \cite{fu2021, sedmdd}, due to the strong correlation between the target phoneme sequence and potential mispronunciation patterns. Transformer-based architectures \cite{wu21h_interspeech} and self-supervised learning (SSL)-based encoders \cite{ryu23_interspeech, fort25_interspeech, w2v2} further strengthened the acoustic backbone and improved robustness, reinforcing the trend that better representations and sequence modeling often translate into better detection and diagnosis.

Beyond architectural improvements, several works refined training strategies and representation learning. Maximum expected F1-score training \cite{9858931} aligned optimization objectives with detection metrics.  \cite{peng2022textawareendtoendmispronunciationdetection, lingw2v2} explicitly addressed mismatches between the learning objectives of phoneme recognition and MDD, as well as dataset imbalance, by enhancing the linguistic encoder branch. Other methods enriched representations by integrating acoustic, phonetic, linguistic, and pitch embeddings \cite{9746604, papl}, modeling distortion errors via anti-phone labels \cite{yan20_interspeech},
adding articulatory features to E2E MDD models \cite{wei25_slate}, or applying focal-attention-based feature fusion with error-type prompts \cite{ZHU2024103009}. These methods demonstrate that improved objectives, richer supervision, and stronger linguistic or acoustic encoders can significantly enhance MDD performance. Multi-view and multi-task learning approaches \cite{elkheir23b_slate, 10448480} incorporated monolingual and multilingual encoders and articulatory feature supervision to enhance phonetic discrimination under low-resource conditions.

Graph-based modeling has been introduced to inject structured priors into pronunciation-related tasks. Hierarchical graph attention modeling has been explored for pronunciation assessment by encoding linguistic hierarchy in heterogeneous graphs \cite{101109TASLP20243449111}. More directly for MDD, graph-based modeling of articulation \cite{10097226} constructs a phonetic graph over the phone inventory and integrates it into a unified dictation-alignment framework. By learning graph-aware prompt embeddings, this approach demonstrates that graph prior knowledge can improve the diagnostic consistency and interpretability of the MDD system.

However, different L1 groups exhibit distinct and systematic phoneme substitution patterns when learning an L2. Modeling all learners with a single, language-agnostic structure can overlook these cross-linguistic differences and weaken diagnostic specificity. Therefore, we are motivated to develop an domain-aware MDD framework that explicitly accounts for language-dependent error tendencies. By separating learners by their L1 backgrounds and constructing language-specific structural priors, the model can better capture systematic confusion dynamics. Moreover, representing these patterns as graphs provides richer relational information than flat labels, since graphs naturally encode connections and directional transitions among commonly confused phonemes.
To that end, we make the following contributions: 
\begin{enumerate}
    \item We construct directed, weighted phoneme confusion graphs for each L1 group using substitution statistics, capturing common and directional patterns of mispronunciation.
    \item We incorporate the graphs into the MDD framework via an L1-specific modeling strategy, thereby enabling the learning of L1-adaptive phoneme representations that encode prior linguistic knowledge.
    \item We demonstrate that our approach achieves superior MDD performance on the L2-ARCTIC benchmark compared to several baselines.
\end{enumerate}

\begin{figure}[htbp]
  \centering
  
  \begin{minipage}{0.7\linewidth}
    \centering
    \includegraphics[width=0.7\linewidth]{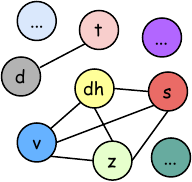}
    \subcaption{Categorical Graph}
    \label{fig:cat_ex}
  \end{minipage}
  
  \vspace{8pt}
  
  \begin{minipage}{0.7\linewidth}
    \centering
    \includegraphics[width=0.7\linewidth]{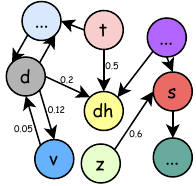}
    \subcaption{Statistical Graph}
    \label{fig:stat_ex}
  \end{minipage}
  
  \caption{Comparison between categorical and statistical phoneme graphs.}
  \label{fig:graph_compare}
\end{figure}

\section{Proposed Method}

\subsection{Graph in MDD}
In MDD, Yan et al.~\cite{10097226} were the first to propose incorporating phone-level articulatory knowledge through a graph-based approach. They observed that phoneme confusions produced by L2 learners often occur within the same articulatory category. For example, /p/ and /b/ are frequently confused because they both belong to the STOP category, and a similar pattern is observed for /s/ and /z/ within the FRICATIVE category. To integrate articulatory traits, they constructed a categorical graph in which all phonemes within the same category are connected, forming a prior knowledge graph that is subsequently incorporated into the MDD model.

However, this method has several limitations. First, certain phonemes from different categories are still frequently confused (e.g., /th/ (FRICATIVE) and /t/ (STOP), or /dh/ (FRICATIVE) and /d/ (STOP)), which cannot be captured by a strictly categorical graph. Second, although some phonemes belong to the same category, they are rarely mispronounced as one another. The categorical graph, however, connects all phonemes within the same category with equal weight, preventing it from modeling the specific influence of one phoneme on another. Finally, the graph is undirected and therefore unable to represent directional mispronunciation patterns. For instance, Vietnamese learners often mispronounce /z/ as /s/, whereas the reverse substitution is less common.

\subsection{Proposed Statistical Graph}
\label{psg}

To overcome the limitations of the categorical graph from prior linguistic assumptions, we construct a statistical phoneme confusion graph directly from the MDD training corpus. Instead of relying on predefined phonological knowledge, our approach quantitatively models learner pronunciation errors by counting observed substitution pairs. Specifically, we collect all mispronunciation instances and compute the frequency with which a produced phoneme is aligned to a target phoneme. This data-driven strategy allows us to identify which phoneme pairs are most frequently confused and to characterize how phonemes are structurally related through learner errors.

First, we extract alignment pairs $(i, j)$ where $i$ represents the canonical phoneme and $j$ represents the learner's realized (mispronounced) phoneme. To focus the graph strictly on error patterns, we exclude correct pronunciations ($i = j$). We define the directed edge weight $w_{j \to i}$ as the conditional probability that phoneme $i$ is realized as phoneme $j$:

\begin{equation}
   w_{j \to i} = \frac{C_{i,j}}{\sum _{k\epsilon V, k\neq i}C_{i,k}}  
\end{equation}

where $C_{i,j}$ denotes the total count of instances where target phoneme $i$ was substituted by phoneme $j$ in the training set, and $\mathcal{V}$ represents the set of all phonemes.

This formulation ensures that the weights of all outgoing error edges from a target phoneme $i$ sum to 1. By structuring the graph this way, the model can learn to propagate phonetic features from frequently substituted error nodes to the target node, thereby capturing the influence of one phoneme on another while effectively modeling the phonetic neighborhood associated with common learner mistakes.

\subsection{MDD via Language-Specific Statistical Graphs (LSSG)}

In MDD, L1 background information is important because there are some typical errors that exist in one language more than the other. Therefore, Yassine et al. \cite{10448480} proposed an L1-aware framework for MDD that incorporates L1 background information into the core MDD system. In detail, L1 background information is integrated into the MDD framework using two approaches. The first is joint training with an additional auxiliary L1 network to predict native language ID, and the second is straightforwardly adding L1 information as a condition before the linear classifier.

\begin{figure*}[h!]
  \centering
    \includegraphics[width=\linewidth]{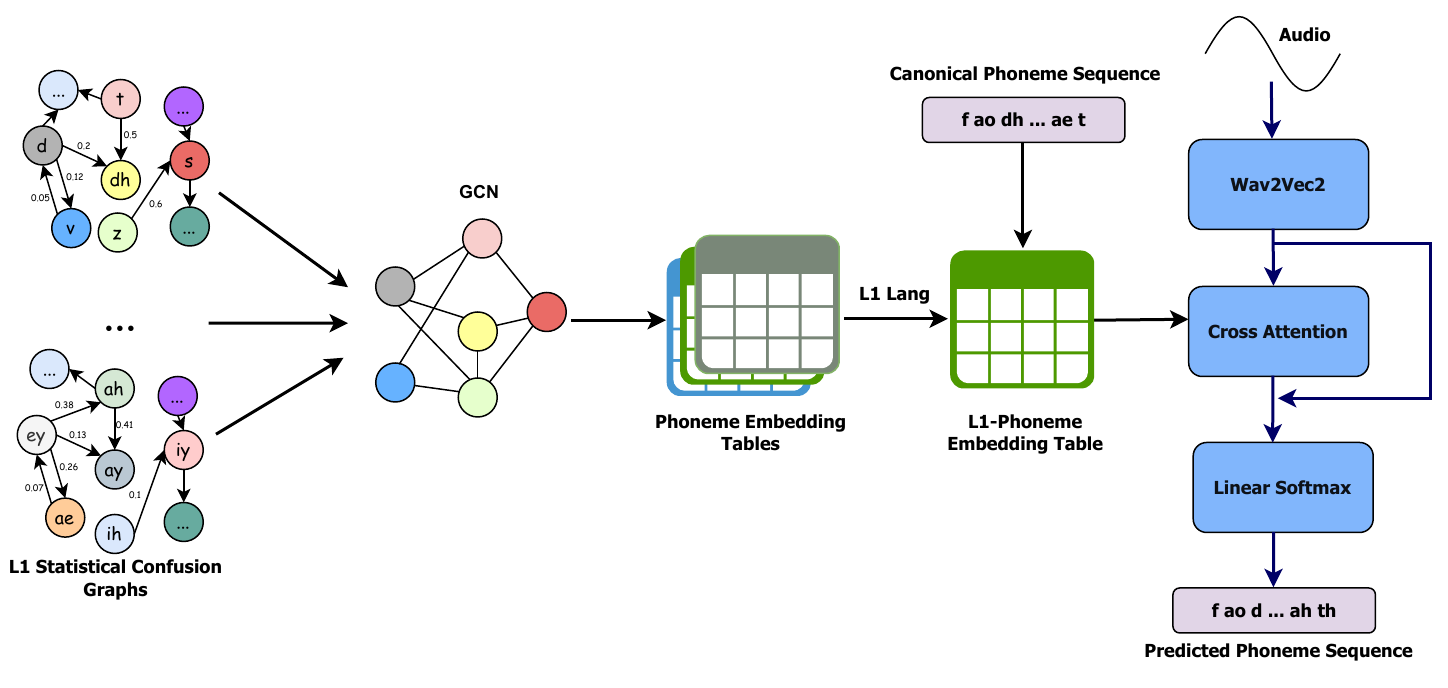}
    \caption{Illustration of our proposed method.}
    \label{fig:pipeline}
\end{figure*}

In contrast to these approaches, in this paper, we incorporate L1 background information into the linguistic branch by leveraging the language-specific statistical graphs constructed in Section~\ref{psg}. Our proposed model, MDD-LSSG, follows a typical MDD framework consisting of an audio encoder and a linguistic encoder, as described in Figure \ref{fig:pipeline}. 

Let $\mathcal{G}^{(l)}$ denote the statistical phoneme confusion graph built from learners with L1 domain $l$. For each mini-batch, given its L1 label $l$, the corresponding graph $\mathcal{G}^{(l)}$ is selected and injected into the linguistic encoder. We employ a Graph Convolutional Network (GCN) \cite{gcn} as a look-up encoder over the phoneme vocabulary $\mathcal{V}$. The L1-dependent phoneme embedding table is computed as:

\begin{equation}
\mathbf{H}^{(l)} = \mathrm{GCN}(\mathcal{V}, \mathcal{G}^{(l)}),
\end{equation}

where the GCN parameters are shared across all languages, and only the edge structure and edge weights of $\mathcal{G}^{(l)}$ vary according to the L1 background.

Given a canonical phoneme sequence $\mathbf{c} = (c_1, c_2, \dots, c_N)$, the linguistic representations are obtained via a look-up operation:

\begin{equation}
\mathbf{L}^{(l)} = \mathbf{H}^{(l)}[\mathbf{c}],
\end{equation}

where $\mathbf{L}^{(l)} \in \mathbb{R}^{N \times d}$. Since $\mathcal{G}^{(l)}$ changes with the language ID, the same canonical sequence may return different linguistic embeddings under different L1 conditions. In this way, L1 background information is encoded structurally through graph propagation rather than through feature-level conditioning.

The L1-adaptive linguistic features $\mathbf{L}^{(l)}$ are then used as keys and values in the cross-attention \cite{attention}, while the acoustic representations $\mathbf{A}$ from the audio encoder serve as queries:

\begin{equation}
\mathbf{C} = \mathrm{CrossAttn}(\mathbf{A}, \mathbf{L}^{(l)}, \mathbf{L}^{(l)}).
\end{equation}

The context vector $\mathbf{C}$ is concatenated with acoustic features $\mathbf{A}$ and forwarded to the linear classifier for CTC-based phoneme prediction.

Through this design, our MDD-LSSG model captures language-dependent pronunciation tendencies by dynamically selecting the statistical graph in the linguistic branch, enabling structurally conditioned phoneme representations.

\section{Experiments}
\subsection{Dataset}

We evaluate our model’s performance using the L2-ARCTIC corpus \cite{zhao18b_interspeech}, a non-native English speech dataset specifically designed for the CAPT task. This corpus includes 24 speakers with L1 backgrounds from various languages, including Arabic, Hindi, Korean, Mandarin, Spanish, and Vietnamese, balanced across genders. It provides canonical phoneme sequences derived from reference texts, corresponding audio recordings read by the speakers, and the actual phoneme transcriptions produced by the speakers, enabling research on the MDD task.

For evaluation, six speakers from different L1 backgrounds, including NJS, TXHC, TLV, ZHAA, YKWK, and TNI, are selected as the test set, while the remaining speakers are used for training.

\subsection{Baseline Models}

To evaluate our approach, we compare our model with several baselines. The first and second baselines follow the L1-aware framework proposed in \cite{10448480}. The first employs an auxiliary network for native language identification alongside the primary acoustic-linguistic MDD network (L1-aware: Aux-Embed). The second baseline incorporates L1 background information as a conditioning input through a lookup embedding, instead of using an auxiliary classification network (L1-aware: Look-up Embed). The third baseline adopts the same backbone architecture as our model but utilizes a categorical graph design similar to that in \cite{10097226} (CAT-GCN-MDD). Finally, we also adopt the MDDGCN model reported in \cite{10097226} as an additional baseline.

\subsection{Experiment Setup}

For the audio encoder, we utilize the wav2vec2-large model \cite{wav2vec2} \texttt{facebook/wav2vec2-large-xlsr-53} from Hugging Face\footnote{\url{https://huggingface.co/facebook/wav2vec2-large-xlsr-53}}. For the graph module, we employ a two-layer GCN with a stacked architecture and residual connections. Node representations are first obtained through an embedding layer, followed by two graph convolution layers that incorporate edge weights from the statistical graph. Residual connections, together with dropout \cite{dropout} and layer normalization \cite{layernorm}, are applied to enhance training stability and facilitate effective information propagation across the graph. The model is trained using the AdamW optimizer \cite{LoshchilovH19} with a batch size of 4, a learning rate of $2e-5$, and a maximum of 100 epochs.

\subsection{Performance Analysis}

\begin{table}
\centering
\caption{Experimental results of different methods on the mispronunciation detection subtask.}
\begin{tblr}{
  cells = {c},
  hlines = {white},
  vlines = {white},
  hline{2} = {-}{black},
  hline{6} = {-}{dotted,black},
}
Models                  & Recall$\uparrow$ & Precision$\uparrow$ & F1$\uparrow$    \\
L1-aware: Aux-Embed \cite{10448480}     & 54.65  & 58.28     & 56.41 \\
L1-aware: Look-up Embed \cite{10448480} & 55.39  & 58.35     & 56.83 \\
MDDGCN \cite{10097226}                  & \textbf{61.67}  & 51.90     & 56.49 \\
CAT-GCN-MDD             & 53.68  & \textbf{63.65}     & \underline{58.24} \\
\textbf{MDD-LSSG (Ours)}                & \underline{57.79}  & \underline{61.36}     & \textbf{59.52} 
\end{tblr}
\label{tab:detection}
\end{table}

\begin{table}
\centering
\caption{Experimental results of different methods on the mispronunciation diagnosis subtask.}

\begin{tblr}{
  cells = {c},
  hlines = {white},
  vlines = {white},
  hline{2} = {-}{black},
  hline{6} = {-}{dotted,black},
}
Models                  & FRR$\downarrow$  & FAR$\downarrow$   & DER$\downarrow$   \\
L1-aware: Aux-Embed \cite{10448480}     & 6.47 & 45.35 & 20.98 \\
L1-aware: Look-up Embed \cite{10448480} & 6.54 & 44.61 & 21.45 \\
MDDGCN \cite{10097226}                  & 9.18 & \textbf{38.03} & 25.24 \\
CAT-GCN-MDD             & \textbf{5.07} & 46.32 & \textbf{20.84} \\
\textbf{MDD-LSSG (Ours)}                & \underline{6.02} & \underline{42.21} & \underline{20.88} 
\end{tblr}
\label{tab:diagnosis}
\end{table}


\begin{figure}[h!]
  \centering
    \includegraphics[width=1\linewidth]{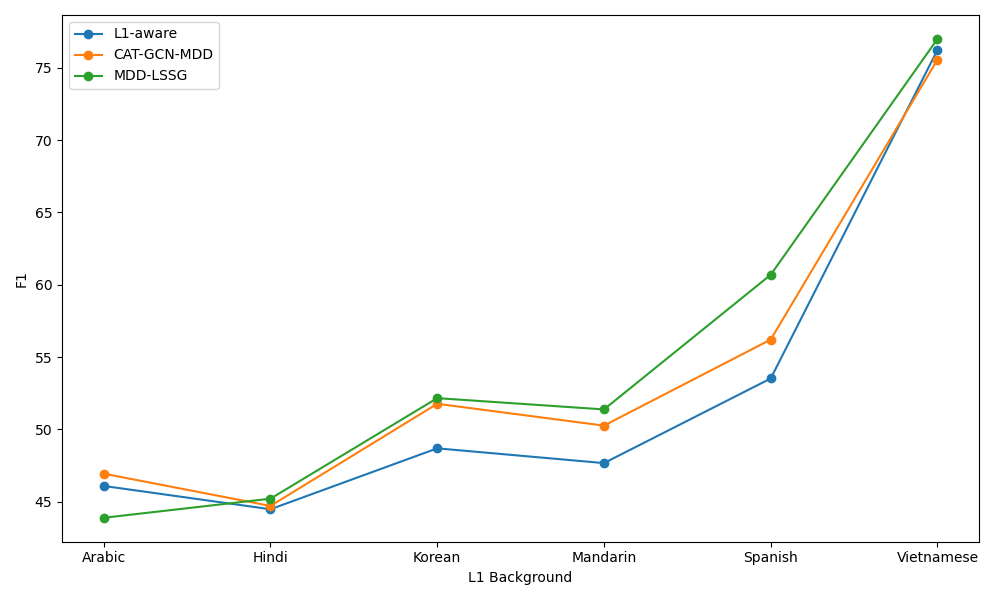}
    \caption{F1-Score Comparison Across L1 Backgrounds}
    \label{fig:f1perlang}
\end{figure}

As outlined in Tables~\ref{tab:detection} and \ref{tab:diagnosis}, our proposed MDD-LSSG demonstrates superior performance in terms of F1-score, surpassing both the L1-aware models as well as MDDGCN and CAT-GCN-MDD, achieving the highest F1-score of 59.52\%. Furthermore, our model exhibits a good balance between Recall and Precision. Regarding the mispronunciation diagnosis subtask, our model also achieves competitive results compared to the baselines.

\begin{figure}[htbp]
  \centering
  
  \begin{minipage}{0.86\linewidth}
    \centering
    \includegraphics[width=0.86\linewidth]{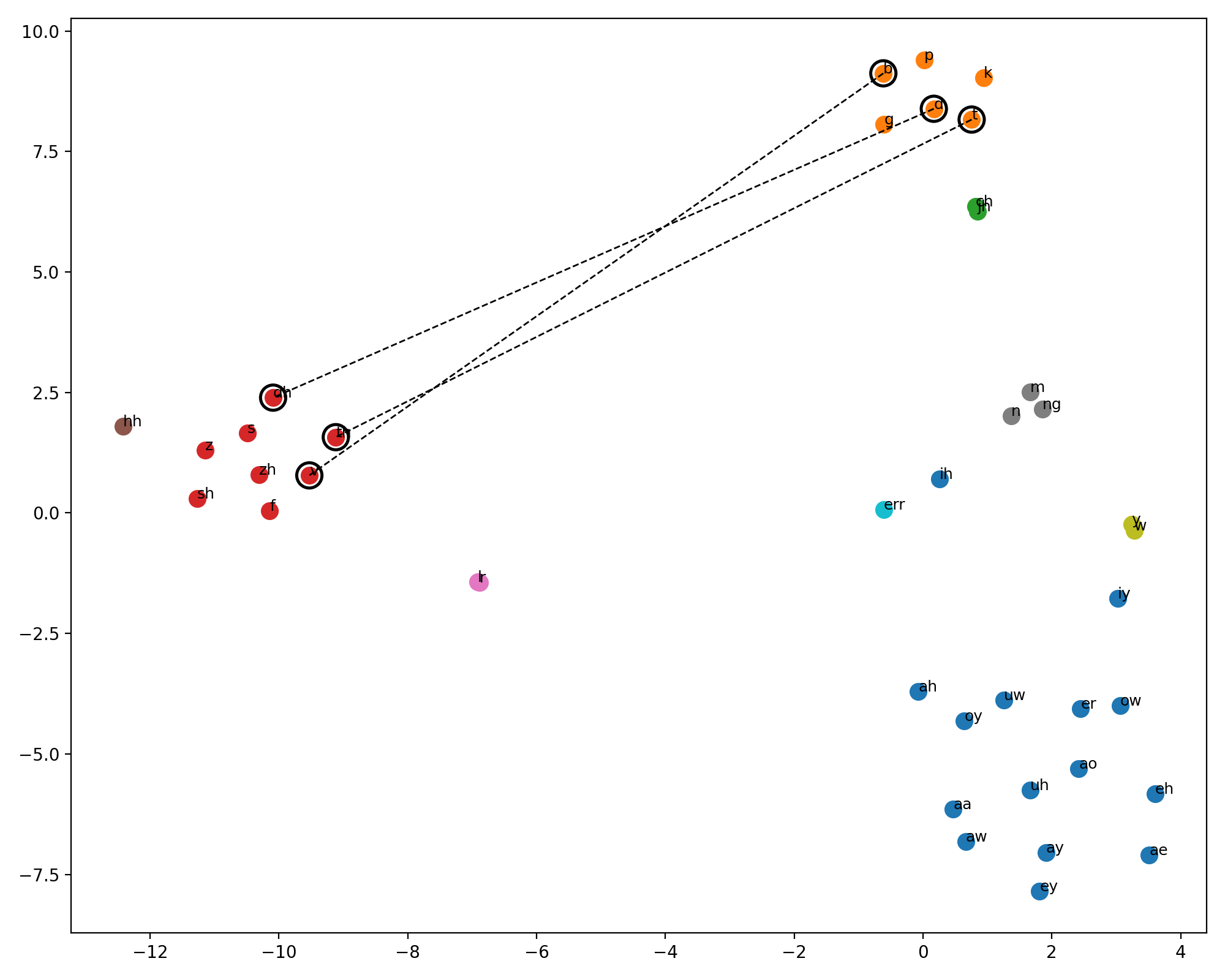}
    \subcaption{Categorical Graph}
    \label{fig:cat_tsne}
  \end{minipage}
  
  \vspace{0.2cm}
  \begin{minipage}{0.86\linewidth}
    \centering
    \includegraphics[width=0.86\linewidth]{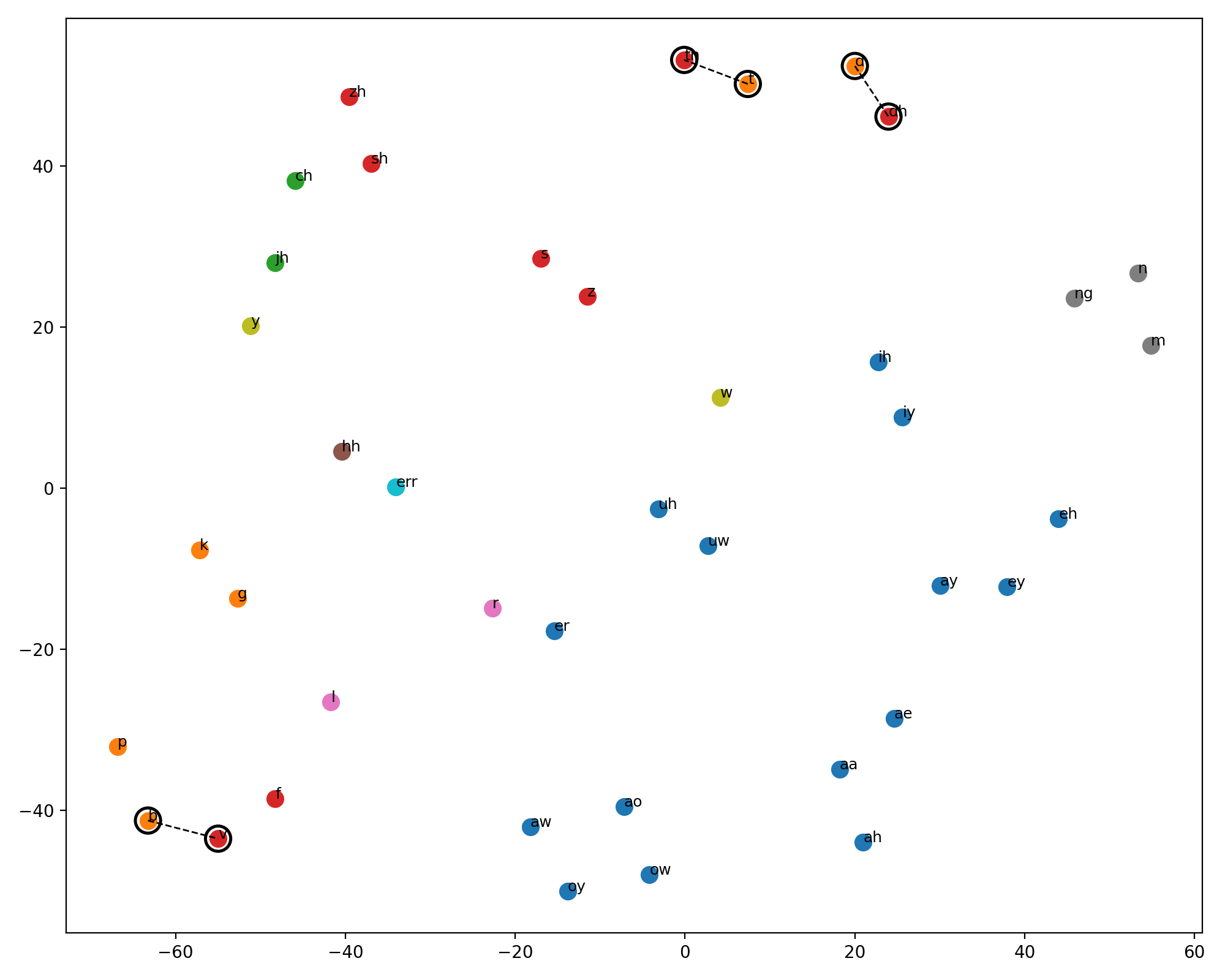}
    \subcaption{Statistical Graph}
    \label{fig:stat_tsne}
  \end{minipage}
  
  \caption{t-SNE Visualization of Phone Embeddings for Spanish}
  \label{fig:graph_compare_tsne}
\end{figure}

We also report the F1-score for each L1 background and compare our model with two baselines: L1-aware: Look-up Embed, which also uses L1 background as a conditioning factor, and CAT-GCN-MDD, which shares the same backbone as our model but employs a categorical graph instead of the proposed statistical graph. As shown in Fig.~\ref{fig:f1perlang}, our model achieves the best F1-score for almost all languages. This emphasizes the importance of using L1-specific graphs, as they lead to improved performance across different L1 backgrounds.

Moreover, we visualize the learned phoneme embeddings using t-SNE, taking Spanish as the L1 background, where our model achieves the highest F1-score and significantly outperforms the baseline. Fig.~\ref{fig:graph_compare_tsne} illustrates the embedding space and highlights several frequently confused phoneme pairs produced by learners. As shown in the figure, although CAT-GCN-MDD forms clear clusters based on phoneme categories, several highly confused pairs for Spanish learners that belong to different clusters, such as /d/-/dh/, /t/-/th/, and /v/-/b/, remain distant in the embedding space. In contrast, our model, which leverages the statistical graph, places these confused phonemes closer together, better reflecting their empirical confusion patterns.

For other pairs that belong to the same category, such as /s/-/z/, /sh/-/zh/, and /ih/-/iy/, which are often mispronounced due to shared phonological properties, our model is still able to preserve meaningful structural relationships, as shown by the close distance between the embeddings of these pairs. These results indicate that the proposed statistical graph captures more informative phoneme representations for canonical phoneme sequence prompts.

\section{Conclusions}

In this paper, we propose a novel method for constructing confusion graphs for the MDD task. Instead of relying on prior linguistic knowledge derived from articulation theory, our approach adopts a data-driven strategy based on quantitative statistical graph modeling. Furthermore, we incorporate language-specific statistical graphs into the MDD framework to better capture L1-dependent pronunciation patterns. The effectiveness of the proposed method is demonstrated through extensive experiments. Our model achieves the best F1-score compared with several recent competitive baselines. In future work, we plan to develop more robust mechanisms to integrate multiple graphs and to further explore different audio encoder backbones in order to enhance overall MDD performance.

\bibliographystyle{IEEEtran}
\bibliography{mybib}

\end{document}